\documentclass[twocolumn]{svjour3}          % twocolumn option for two-column layout

\usepackage{graphicx}                       % standard LaTeX graphics tool
\usepackage{amsmath,amsfonts,amssymb}       % for mathematical symbols and fonts
\usepackage{hyperref}                       % for hyperlinks
\usepackage{float}                          % for placing figures/tables
\usepackage{lipsum}                         % for dummy text (optional)
\usepackage{newtxtext}                      % Times font for text
\usepackage{newtxmath}                      % Times font for math (with bold support)
\usepackage{multirow}                       % for multi-row tables
\usepackage[title]{appendix}                % for appendices
\usepackage{xcolor}                         % for color
\usepackage{textcomp}                       % for text symbols
\usepackage{booktabs}                       % for better table rules
\usepackage{algorithm}                      % for algorithm environment
\usepackage{algpseudocode}                  % for pseudocode within algorithms
\usepackage{listings}                       % for code listings
\usepackage{ctable}
\usepackage[left, pagewise, switch]{lineno}  % Line 
\usepackage[square,numbers]{natbib}

\journalname{Springer Journal}              % Define the journal name (optional)

\title{eXpLogic: Explaining Logic Types and Patterns in DiffLogic Networks}

\author{Stephen Wormald \textsuperscript{1*} \and 
        David Koblah\textsuperscript{1} \and 
        Matheus Kunzler Maldaner\textsuperscript{1} \and 
        Domenic Forte\textsuperscript{1} \and 
        Damon L. Woodard\textsuperscript{1}}
\institute{
    \textsuperscript{1}Florida Institute for National Security, University of Florida, 601 Gale Lemerand Dr, Gainesville, FL 32611, Gainesville, 32611, Florida, United States. \\
    *Corresponding author(s). E-mail(s): \href{mailto:stephen.wormald@ufl.edu}{stephen.wormald@ufl.edu} \\
    Contributing authors: \href{mailto:mkunzlermaldaner@ufl.edu}{mkunzlermaldaner@ufl.edu}; \href{mailto:dkoblah@ufl.edu}{dkoblah@ufl.edu}; \href{mailto:dforte@ece.ufl.edu}{dforte@ece.ufl.edu}; \href{mailto:dwoodard@ece.ufl.edu}{dwoodard@ece.ufl.edu};\\
}

\newcolumntype{C}[1]{>{\centering\arraybackslash}p{#1}}

\date{}
\begin{document}

\maketitle

\begin{abstract}
Constraining deep neural networks (DNNs) to learn individual logic types per node, as performed using the DiffLogic network architecture, opens the door to model-specific explanation techniques that quell the complexity inherent to DNNs. Inspired by principles of circuit analysis from computer engineering, this work presents an algorithm (eXpLogic) for producing saliency maps which explain input patterns that activate certain functions. The eXpLogic explanations: (1) show the exact set of inputs responsible for a decision, which helps interpret false negative and false positive predictions, (2) highlight common input patterns that activate certain outputs, and (3) help reduce the network size to improve class-specific inference. To evaluate the eXpLogic saliency map, we introduce a metric that quantifies how much an input changes before switching a model's class prediction (the SwitchDist) and use this metric to compare eXpLogic against the Vanilla Gradients (VG) and Integrated Gradient (IG) methods. Generally, we show that eXpLogic saliency maps are better at predicting which inputs will change the class score. These maps help reduce the network size and inference times by 87\% and 8\%, respectively, while having a limited impact (-3.8\%) on class-specific predictions. The broader value of this work to machine learning is in demonstrating how certain DNN architectures promote explainability, which is relevant to healthcare, defense, and law. 
\end{abstract}

\section{Introduction}
\label{introduction}

\begin{figure}[htp!]
\centering
\includegraphics[width=6cm, trim={8cm 7.5cm 8.5cm 5cm},clip]{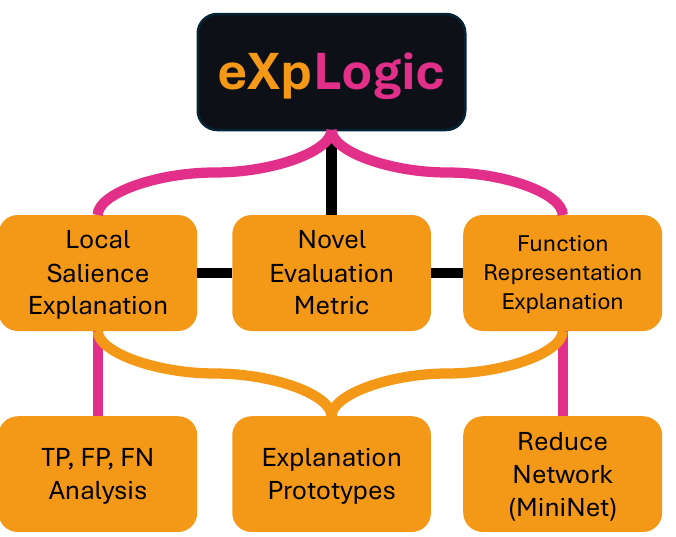}
\caption{Overview of several explanation tools that are enabled when using the DiffLogic network, which are referred to as the eXpLogic set of analyses}
\label{fig:toolset}
\end{figure}

Understanding the behaviors of DNNs could aid the reliable adoption of machine learning in fields like healthcare, defense, and law, where interpretability and trust are paramount \citep{tiwari2023explainable}. The field of eXplainable Artificial Intelligence (XAI) approaches this need by providing post-hoc methods for interpreting black-box models, and by designing architectures that are inherently more interpretable \cite{adadi2018peeking}. The DiffLogic network presented by Petersen et al. \cite{petersen2022deep} is a unique architecture that aids interpretability -- each node learns to represent one of sixteen logic types (e.g. AND, OR, etc. per Table \ref{tab:gates}). While DiffLogic makes it possible to interpret the logical behaviors of internal nodes, the original authors do not leverage this trait to explain their model's behaviors. This paper introduces eXpLogic to fill this research gap (seen in Figure \ref{fig:toolset}). 

\begin{table}[!ht]
    \caption{Table of logic gates learned by the DiffLogic network showing the probability of a gate's output ($Z$) given two inputs A and B} 
    \scriptsize
    \centering
    \begin{tabular}{cccc}
    \toprule
        ID & AB:(00,01,10,11) & Logic Gate & P($Z$) \\ \hline
        
        0  & (0, 0, 0, 0) & FALSE & 0 \\ 
        1  & (0, 0, 0, 1) & AND(A,B)  & P(A)P(B) \\ 
        2  & (0, 0, 1, 0) & AND(A,NOT(B))  & P(A) - P(A)P(B) \\ 
        3  & (0, 0, 1, 1) & A & P(A) \\ 
        4  & (0, 1, 0, 0) & AND(NOT(A),B)  & P(B) - P(A)P(B) \\ 
        5  & (0, 1, 0, 1) & B & P(B) \\ 
        6  & (0, 1, 1, 0) & XOR(A,B)  & P(A) + P(B) - 2*P(A)P(B) \\ 
        7  & (0, 1, 1, 1) & OR(A,B)  & P(A) + P(B) - P(A)P(B) \\ 
        8  & (1, 0, 0, 0) & NOR(A,B)  & 1 - (P(A) + P(B) - P(A)P(B)) \\ 
        9  & (1, 0, 0, 1) & XNOR(A,B)  & 1 - (P(A) + P(B) - 2*P(A)P(B)) \\ 
        10 & (1, 0, 1, 0) & NOT(B) & 1 - P(B) \\ 
        11 & (1, 0, 1, 1) & OR(A,NOT(B)) & 1 - P(B) + P(A)P(B) \\ 
        12 & (1, 1, 0, 0) & NOT(A) & 1 - P(A) \\ 
        13 & (1, 1, 0, 1) & OR(NOT(A), B) & 1 - P(A) + P(A)P(B) \\ 
        14 & (1, 1, 1, 0) & NAND(A,B)  & 1 - P(A)P(B) \\ 
        15 & (1, 1, 1, 1) & TRUE & 1 \\ \hline 
                
    \end{tabular}
    \label{tab:gates}
\end{table}

When applied to DiffLogic models trained for classification, eXpLogic highlights the importance of certain inputs in several scenarios. Specifically, eXpLogic: (i) Reveals which inputs caused a given class prediction (i.e., a local saliency method); (ii) Shows which inputs are likely to ``turn on" a selected function, which helps explain associated input patterns (i.e. a function explanation method); (iii) Helps model inspectors understand True Positive (TP), False Positive (FP) and False Negative (FN) predictions by providing comparable evidence for each case; and (iv) Produces function explanations that show which inputs and networks nodes can be ignored, making it possible to prune irrelevant functions from DiffLogic networks to create miniature networks (MiniNets) that predict individual classes with reduced inference times. While many methods produce saliency maps \cite{samuel2021evaluation} and reveal input patterns that activate certain functions \cite{erhan2009visualizing}, eXpLogic differs by revealing explicate sets of important inputs by leveraging the logical structure learned by DiffLogic. We introduce a novel explanation metric to compare the eXpLogic method against existing methods.

To present this work, Section \ref{sec:background} demonstrates the novelty of eXpLogic and Section \ref{sec:methods} outlines the methodology. Section \ref{sec:results} illustrates results from the explanation method, compares eXpLogic against existing approaches, and studies how MiniNets increase class-specific inference speed. Salient conclusions are presented in Section \ref{sec:conclusion}. 

\section{Background and Novelty}
\label{sec:background}

DiffLogic, or ``Differentiable Logic Gate Networks," is a DNN architecture that constrains each node in the DNN to represent real-logic operations~\cite{petersen2022deep}. DiffLogic can be interpreted as a set of logic gates, making it generally more interpretable than standard multi-layer-perceptrons (MLP)." Since its publication, the DiffLogic architecture has gained attention in explainable convolutional neural networks \cite{benamira2022scalable} and concept reasoning \cite{vemuri2024enhancing}. The current work extends existing research in several key ways, as described through the following list of contributions:  

\begin{itemize}

    %\item \textit{Novel Extension of DiffLogic Focused on Explanation}: While DiffLogic gains inherent interpretability due to its reliance on logic networks, existing works take the interpretability for granted when designing explainable convolutional neural networks \cite{benamira2022scalable} and concept reasoners \cite{vemuri2024enhancing, barbiero2023interpretable} and do not focus on more deeply explaining behaviors learned by DiffLogic in human-comprehensible formats. This paper is the first known work to leverage the logic learned by the DiffLogic for the benefit of explaining model behaviors. 

    \item \textit{Novel Input Saliency Explanation}: While many saliency methods exist, including GradCAM, SHAP, and DeepLift \cite{samuel2021evaluation}, they can give misleading information \cite{boccignone2019problems} that are preferred for visual clarity rather than their sensitivity to model behaviors \cite{adebayo2018sanity}. The local eXpLogic saliency method (eXpLogic-L) is novel for showing the exact set of inputs that are responsible for a given decision (in the case of binary inputs), and for revealing whether an input's presence or absence supports a model's decision. 

    \item \textit{Novel Function Explanation}: Nodes in a DNN can be explained by showing the input patterns that activate certain functions. A comparable method called activation maximization \cite{erhan2009visualizing} produces example inputs via gradient ascent, which is prone to finding local maxima. The eXpLogic method for explaining a function exploits the observation that DiffLogic learns a limited set of decision traces, which can be used to identify and weight the full set of inputs associated with certain functions according to the likelihood of each decision trace. To the authors knowledge, this method is a novel approach of interpreting DiffLogic networks.     
    
    \item \textit{Novel Metric for Evaluating Saliency Methods}: Many metrics exist to evaluate saliency methods. Some compare saliency maps against human eye-fixations \cite{riche2013saliency}, others measure model sensitivity, human perceptibility, and performance characteristics \cite{boggust2023saliency}. When evaluating a model's sensitivity, the Area Over the Perturbation Curve (AOPC) metric \cite{tomsett2020sanity} quantifies how much a model output changes as the current input is varied to some baseline. However, when classification scores returned by a model are discrete, the AOPC sums to zero until the class changes. To account for this issue, we introduce the SwitchDist (the distance needed to switch a model's class prediction) as a metric to evaluate saliency methods in discrete classification settings. 

\end{itemize}

In summary, eXpLogic is a single algorithm which produces local explanations of important inputs, and function explanations of input patterns that result in logical highs for studied functions. The novelty of this work and forthcoming applications are realized via the methodology discussed in the following section.

\section{Methods}
\label{sec:methods}

The eXpLogic method uses a single algorithm (Section \ref{sec:method:algorithm}) to highlight inputs that provides evidence for TP, FN, and FP predictions, and to reveal inputs associated with individual functions (Section \ref{sec:method:local}). These explanations are compared against existing saliency methods (Section \ref{sec:method:eval}) before demonstrating that eXpLogic helps prune DiffLogic networks and produce MiniNets that reduce the inference time needed to predict individual classes (Section \ref{sec:method:inference}). 

\subsection{Algorithm for Local and Function Explanations}
\label{sec:method:algorithm} 

The eXpLogic algorithm leverages the observation that DiffLogic learns aspects of a model's structure by implicitly pruning the ``activation paths" being used to classify certain outputs. Here, an activation path represents a single path through the learned logic gates, from input to output \cite{jiang2023explaining}, and are ranked by importance using several metrics below. In contrast to MLPs, each logic gate learned by DiffLogic has a maximum of two inputs such that a node on layer L can be connected to at most $2^L$ inputs. Because a network trained to classify $C$ classes using N nodes per layer predicts class $C_i$ by summing $(N/|C|)$ from the second to last layer, class $C_i$ is connected to at most $(2^L)(N/|C|)$ activation paths through the network. However, by learning the TRUE or FALSE gate types (ID 0 and 15 from Table \ref{tab:gates}), the path is effectively blocked. Furthermore, the gate types learned along an activated path determines whether a connected input activates downstream classes when it is present (closer to one) or absent (closer to zero). We use this observation to structure Algorithm \ref{alg:scaled_fanin}, which performs a breadth-first search over all activation paths upstream of a selected function (representing that functions ``fan in," or $FANIN$) while tracking which paths pass through NOT gates. The result is the set of input dimensions (or $I$) that influence the selected input either in their presence (when 1) or absence (when 0). Here, the ``$SumSigns(I)$" function sums the signs associated with each input and formats $I$ for visualization.   

\begin{algorithm}
\caption{ - FanIn($DL$, $v_{out}$, $\theta$): Find $v_{out}$ Fan-in}
\begin{algorithmic}[1]
\Require DL Net. $(DL)$, DL node $v_{out}$, threshold $\theta$
\State $Q \gets \{(v_{out}, 1)\}$ \Comment{Queue of nodes to check}
\While{$Q$ is not empty}
    \State $(v, s) \gets \texttt{Dequeue}(Q)$
    \For{each predecessor $u$ of $v$ in $DL$}
        \State $s_{new} \gets s$
        \State $g_{type} \gets \texttt{GateType}(v)$
        \State ($G_{not-list}) \gets \texttt{EdgeType}(u, v)$
        \If{$(g_{type} \in G_{not-list})$}
            \State $s_{new} \gets -s_{new}$
        \EndIf
        \State $\theta_{comp} \gets \begin{cases} \theta & \text{if } s_{new} = 1 \\ 1 - \theta & \text{if } s_{new} = -1 \end{cases}$
        \If{$SaliencyFactor(u) > \theta_{comp}$}
            \State $\texttt{Enqueue}(Q, (u, s_{new}))$
        \ElsIf {$g_{type}$ = $``input"$}
            \State $I \gets I \cup \{(u, s_{new})\}$
        \EndIf
    \EndFor
\EndWhile
\State $I_{Saliency} \gets \texttt{SumSigns}(I)$ 
\\
\Return $I_{Saliency}$
\end{algorithmic}
\label{alg:scaled_fanin}
\end{algorithm}

Note Algorithm \ref{alg:scaled_fanin} includes a threshold $\theta$, which prunes activation paths from being included based on a ``SaliencyFactor(u)" metric, refereed to as ``\texttt{SF}." Numerous metrics can be used, including: (i) The Empirical Signal Probability per Equation \eqref{eq:SP_E}, or \texttt{$\texttt{SP}_E$}, and (ii) the Analytical Signal Probability per Equation \eqref{eq:SP_A}, or \texttt{$\texttt{SP}_A$}. Here, $\texttt{SP}_E$ is an empirical metric representing how often a logic gate's output ($Z$) is active (produces a logic high) given a set of inputs ($X$). $\texttt{SP}_A$ is an analytical variant derived from the real logic expressions of each gate type, which are summarized in Table \ref{tab:gates}. Each metric is calculated per layer ($L$) and node ($n$) in a DiffLogic network (DL), as indicated by the subscripts below: 

\begin{equation}\label{eq:SP_E}
    \texttt{SP}_{E,(L,n)}(\textbf{X}) = \frac{\sum_{i=1}^{|\textbf{X}|}DL_{(L,n)}(\textbf{X}_i)}{|\textbf{X}|}
\end{equation}

\begin{equation}\label{eq:SP_A}
    \texttt{SP}_{A,(L,n)}(\textbf{X}) = P_{(L,n)}(Z|\textbf{X})
\end{equation}

After calculating a given \texttt{SF} per node, a set of saliency maps can be generated using $FanIn()$ across a range of $\theta$ and averaged using Equation \eqref{eq:integrate} to show which inputs are associated with common activation paths, as reported in later sections. 

\begin{equation}\label{eq:integrate}
    eXpLogic_\texttt{SF} = \frac{1}{\theta_{max} - \theta_{min}}\sum_{\theta=\theta_{min}}^{\theta_{max}} FanIn_{\texttt{SF}}(DL, n, \theta) 
\end{equation}

\subsection{Training DiffLogic Networks}
\label{sec:method:training} 

This work iteratively trained a set of DiffLogic models while reducing the number of layers and nodes so the test accuracy exceeded 90\% on a binarized version of the MNIST dataset. MNIST was selected given its prevalence in benchmarking saliency methods. The images were cropped to 20x20 pixels to remove the blank boarder, and models were trained using a train-test split of 80\%-20\%, a learning rate of 0.01, a ``grad-factor"\footnote{The ``grad-factor" defined in DiffLogic avoids vanishing gradients} of 1, and a $\tau$ of 10. The resulting network had two layers with 2500 nodes. Five models of these models were trained to demonstrate eXpLogic, achieving a mean test accuracy of 91.2\%. While DiffLogic can achieve test state-of-the-art performance, this work sought a small network to aid explainability and to help determine whether the eXpLogic method is useful for understanding FN and FP predictions (shown in Section \ref{sec:method:local}). 

\subsection{Demonstrating eXpLogic Explanations}
\label{sec:method:local} 

The eXpLogic algorithm (Algorithm \ref{alg:scaled_fanin}) was applied to explain \textit{function representations} and \textit{local predictions} as summarized in Table \ref{tab:types}. While local predictions are explained using single input $x_i$, function behaviors are explained using both $\mathcal{D}$, subsets of data per class type ($\mathcal{D}_{C_i}$), the null set ($\emptyset$), and for a uniform input probability of 50\% ($\mathcal{U}$). The \texttt{SF} metrics were used to produce ten explanation types, which are categorized as local (eXpLogic-L), global (eXpLogic-G), and class (eXpLogic-C)  explanations. Using Algorithm \ref{alg:scaled_fanin} to explain the five models, eXpLogic-L explanations were generated for each sample in the MNIST dataset, while the alternate types were used to explain each output node in the DiffLogic network. The eXpLogic-L saliency maps for TP, FN, and FP predictions demonstrate how eXpLogic helps model inspectors understand incorrect model classifications, which helps in debugging model behaviors \cite{yona2021revisiting}. All explanations (excluding eXpLogic-$G_\emptyset$\footnote{eXpLogic-$G_\emptyset$ is used for visualization purposes}) are evaluated per the following methods.

\subsection{SwitchDist Metric to Evaluate eXpLogic Explanations}
\label{sec:method:eval} 

This section introduces a novel metric for comparing the local and function explanations against comparable saliency maps produced by the VG and IG methods \cite{sundararajan2017axiomatic}. In general, saliency methods reveal which inputs are important for predicting the current class, or what inputs may be changed to alter the class score. Each saliency method implicitly reveals one direction in the image space that could change the class score (see Figure \ref{fig:saliency}.c), such as the VG method which shows the gradient of the class score. However, directions beside the gradient could change the class score more quickly. Therefore, we propose to rank saliency methods by the distance needed to switch a model's class prediction, or the ``SwitchDist," which applies to continuous and discrete model outputs. In this work, each saliency method produces a saliency map $\mathbf{S}$ which corresponds to some unit vector in the image space, or $\mathbf{\hat{U}} = unit(\mathbf{S})$, as defined per Equation \eqref{eq:unit}.

\begin{equation} \label{eq:unit}
    \text{unit}(\mathbf{X}) = \frac{\mathbf{X}}{\|\mathbf{X}\|_2}
\end{equation}

 $\mathbf{\hat{U}}$ is used to define three directions in the image space per $\mathbf{S}$ that either remove important positive values (+I), add important negative values (-I), or do both (+/-I). Images were updated per the equations in Table \ref{tab:directions} using a step size (or $\alpha$) of 0.1 until the model's class score switched, or until $\|x'_t - x_i\|_2 > |x_i|/4$. Section \ref{sec:results} summarizes results from characterizing the SwitchDist.   

\begin{table}[!ht]
    \caption{eXpLogic saliency map variations which higlight different activation paths. Different inputs and the range of $\theta$ are used for each} % Add a caption if needed
    \label{tab:types}
    \scriptsize
    \centering
    \begin{tabular}{c c c c}
    \hline
        Variant      & Metric & [$\theta_{min}$; $\Delta\theta$; $\theta_{max}$]     \\ 
        \hline 
        eXpLogic-$L_E$ & $\texttt{SP}_E$($x_i$)&       [0; 0; 0] \\ 
        eXpLogic-$L_A$ & $\texttt{SP}_A$($P(x|x_i)$)&  [0; 0; 0] \\ 
        eXpLogic-$G_\emptyset$   & $\texttt{SP}_E$($\emptyset$) & [0; 0.01, 1] \\ 
        eXpLogic-$G_\mathcal{U}$ & $\texttt{SP}_A$($P(x|\mathcal{U})$) & [0; 0.01, 1] \\  
        eXpLogic-$G_E$ & $\texttt{SP}_E$($\mathcal{D}$) & [0; 0.01; 1] \\ 
        eXpLogic-$G_A$ & $\texttt{SP}_A$($P(x|\mathcal{D})$) & [0; 0.01; 1] \\  
        eXpLogic-$C_E$ & $\texttt{SP}_E$($\mathcal{D}_{C_i}$) & [0; 0.01; 1] \\ 
        eXpLogic-$C_A$ & $\texttt{SP}_A$($P(x|\mathcal{D}_{C_i})$) & [0; 0.01; 1]\\ 
        \hline 
 
    \end{tabular}
\end{table}

\begin{table}[h!]
\centering
\scriptsize
\caption{Three unit directions derived from a saliency map for altering a model's input, where the tolerance $\delta$ is set to the standard deviation of $(\mathbf{U} - \mu_\mathbf{U})/100$, where $\mu_\mathbf{U}$ is the mean of $\mathbf{U}$}
\begin{tabular}{ c c c}
\hline
\textbf{Symbol}      & \textbf{Method}     & \textbf{Equation} \\ \hline  
+/-I             & Switch All Imp.     & \( x' = x - \alpha \cdot \mathbf{U} \) \\ 
+I               & Remove (+) Imp.       & \( x' = x - \alpha \cdot \text{unit}(\mathbf{U} \cdot (\mathbf{U} + \delta > 0)) \) \\ 
-I               & Add (-) Imp.       & \( x' = x - \alpha \cdot \text{unit}(\mathbf{U} \cdot (\mathbf{U} + \delta <= 0)) \) \\ \hline 
%+/- (!) Imp.         & Add All ! Imp.      & \( x' = x + \alpha \cdot \text{Norm}(\mathbf{U} \cdot \text{!\texttt{Imp.}}) \) \\ \hline
%+ (!) Imp.           & Add + ! Imp.        & \( x' = x + \alpha \cdot \text{Norm}(\mathbf{U} \cdot \text{!\texttt{Imp.}} \cdot \mathbb{I}(x = 1)) \) \\ \hline
%- (!) Imp.           & Add - ! Imp.        & \( x' = x + \alpha \cdot \text{Norm}(\mathbf{U} \cdot \text{!\texttt{Imp.}} \cdot \mathbb{I}(x = 0)) \) \\ \hline
\end{tabular}
\label{tab:directions}
\end{table}

\subsection{Reducing Model Inference Speeds}
\label{sec:method:inference} 

The $FANIN$ found using eXpLogic can also be used to reduce model size and resulting inference speeds. While the class-specific features learned by standard DNNs can be hard to disentangle, the $FANIN$ for an output node in a DiffLogic network represents the subset of features and functions relevant to predicting an individual class. Ten class-specific miniature networks (or MiniNets) were created by retaining only the nodes in a class's $FANIN$. While DiffLogic predicts a class by taking the argmax of the class-sum, each MiniNet has a single output so maximum likelihood estimation was performed on the True and False MiniNet predictions to determine an appropriate decision threshold. Each Mininet was compared against the parent network using the accuracy, relative network size, and relative inference time, as summarized in the following section.

\section{Results and Discussion}
\label{sec:results}
The following sections show and evaluate the local and function explanations produced by eXpLogic (Section \ref{sec:results:explanations}) before revealing the value of Mininets in inference (Section \ref{sec:results:inference}).  

\begin{figure*}[h!]
\centering
\includegraphics[width=\textwidth, trim={2cm 7.3cm 1cm 6cm},clip]{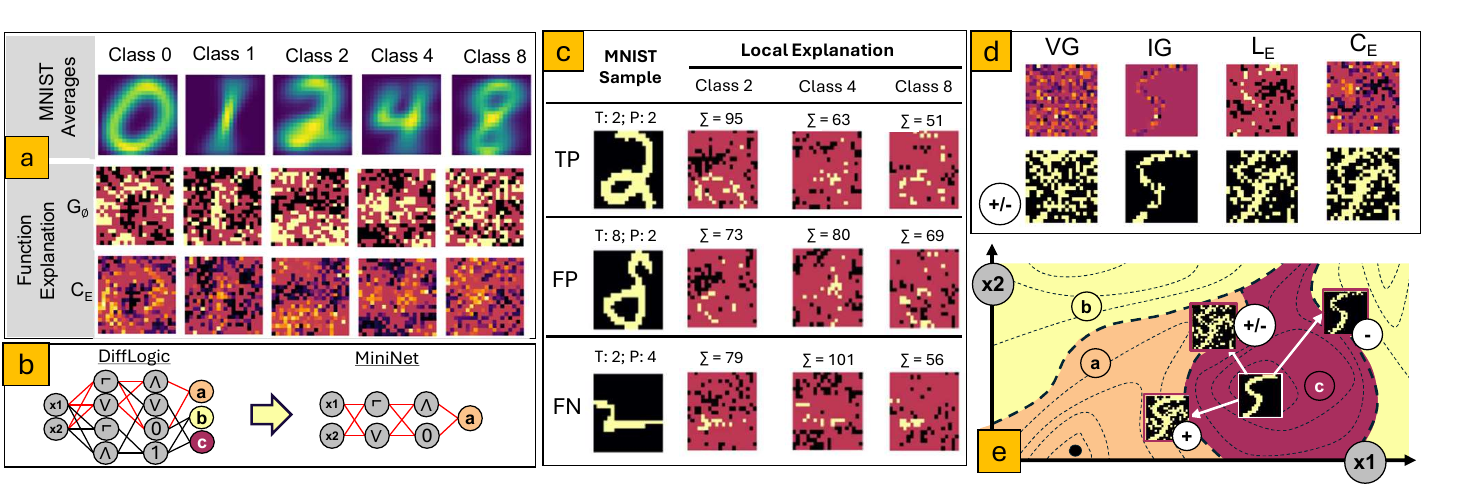}
\caption{Overview of eXpLogic, showing: (a) examples of the eXpLogic function explanations, where each exhibits input patterns which influence each digit class from the MNIST dataset; (b) a notional \texttt{DiffLogic} architecture that predicts three classes which is juxtapozed with a MiniNet derived from the FANIN of ``class a;" (c) Local explanation method showing the evidence (important pixels) that support the decisions of class 2, 4, and 8 from the MNIST dataset. Each row shows either a TP, FP, or FN prediction where the image titles show the True (T) and Predicted (P) labels for the MNIST images, and the number of important inputs ($\Sigma$) identified for each explanation. Note that $\Sigma$ is not always highest for the predicted class. Light pixels represent important positive inputs, whereas dark pixels represent important zero inputs; (d) a graphical illustration of the SwithDist used to evaluate each saliency methods, where a current input, class 5 in this case, is modified in (e) three directions which either add, remove, or alter the important input pixels. Each image represents a baseline of class 5 modified per each unit direction from Table \ref{tab:directions}}
\label{fig:saliency}
\end{figure*}

\subsection{Results from Explaining DiffLogic Behaviors} 
\label{sec:results:explanations} 

Function and local explanations illuminate diverse model behaviors, which may both be evaluated using the SwithDist as described in the following bullet points:  

\begin{itemize}
    \item \textit{Examples of Function Explanations}: Patterns in the MNIST dataset are evident in the eXpLogic explanations. Figure \ref{fig:saliency} shows the function explanations associated with each class and indicate that each output node has learned a preference for certain inputs. The pixel colors indicate whether a class activates in response to the presence or absence of certain inputs. For example, eXpLogic-{$G_\emptyset$} explanations show whether a class is likely to be predicted when there are positive values (bright pixels), or zero values (dark pixels) in certain input dimensions. While some explanations look like the class--averages from MNIST -- like classes zero and one - others reveal a preference for spatial regions which are distinct to a certain class -- like classes two and four, which exhibit a preference for positive inputs in the bottom and center regions of the image, respectively. Each eXpLogic explanation type generally highlights similar inputs, but with varying intensities depending on the \texttt{SF} metric used and the resulting activation paths found as important when using Algorithm \ref{alg:scaled_fanin}.
    \item \textit{Examples of Local Explanations}: While function explanations represent the sets of inputs that \textit{could} predict each class, the local explanations in Figure \ref{fig:saliency}.c show which inputs support individual predictions. In the TP case for class 2, a set of 95 active and non-active inputs supported the class decision. The FP case shows how a dark center present in an image from class 8 supports the false prediction of class 2. For the FN case, a horizontal line is next to dark pixels in the top and bottom of the images, which results in an image of a ``2" being labeled a 4. These examples further demonstrate the unique ability of eXpLogic to reveal when the absence or presence of certain inputs support a class decision, which is uncommon in alternate saliency methods.
    \item \textit{Evaluating Local and Function Explanations}: Table \ref{tab:switchdist} evaluates the SwitchDist associated with each saliency method. In general, the pixels highlighted by eXpLogic changed the class score more quickly than pixels marked by the VG and IG methods. In this analysis, VG and IG often perform worse than a random baseline. While altering the important zero values using eXpLogic saliency maps often resulted in low SwithDist values, removing only the important positive pixels produced worse SwithDist values. These results could indicate that eXpLogic is better at predicting important negative values, or that the trained network relies heavily on negative values to make predictions. 
\end{itemize}

 Together, these results seem to indicate that both local and function explanations produced by eXpLogic perform better than VG and IG baselines, which aids model interpretability.

%\begin{figure}[h]
%\centering
%\includegraphics[width=7cm, trim={7.5cm 4.75cm 7.5cm 5cm},clip]{Figure_x2.pdf}
%\caption{Local explanation method showing the evidence (important pixels) that support the decisions of class 2, 4, and 8 from the MNIST dataset. Each row shows either a TP, FP, or FN prediction where the image titles show the True (T) and Predicted (P) labels for the MNIST images, and the number of important inputs ($\Sigma$) identified for each explanation. Note that $\Sigma$ is not always highest for the predicted class. Light pixels represent important positive inputs, whereas dark pixels represent important zero inputs}
%\label{fig:local}
%\end{figure}

\begin{table*}[htp!]
    \centering
    \caption{Comparison of the SwitchDist per saliency type and derived unit direction from Table \ref{tab:directions} produced for each image predicted as TP from MNIST for each of the five trained models. The three lowest values of the SwithDist are bolded per unit direction. Not all images switched class predictions when altering pixels, as represented using the \%Switch metric that shows how many images switched classes for an explanation type}
    \small
    \begin{tabular}{C{3.0cm}C{2.25cm}C{1.75cm}C{2.25cm}C{1.75cm}C{2.25cm}C{1.75cm}}
        \hline
        & \multicolumn{2}{c}{Remove (+/-) Importance} & \multicolumn{2}{c}{Remove (+) Importance} & \multicolumn{2}{c}{Remove (-) Importance} \\
        \hline
        Method & SwitchDist ($L_2$) & \% Switch & SwitchDist ($L_2$) & \% Switch & SwitchDist ($L_2$) & \% Switch \\
        \hline 
        Random          & 7.9 $\pm$ 2.6   &  (92.1\%) & \textbf{7.9 $\pm$ 2.6}   &  (92.1\%)  & 7.9 $\pm$ 2.6  &  (92.1\%) \\
        Vanilla Gradients              & 12.8 $\pm$ 11.4 &  (69.5\%) & 12.7 $\pm$ 13.8 &  (15.6\%)  & 9.7 $\pm$ 11.2 &  (53.5\%) \\ 
        Integrated Gradients              & 12.8 $\pm$ 15.1 &  (18.6\%) & \textbf{10.6 $\pm$ 14.6} &  (18.9\%)  & NaN $\pm$ NaN  &  (0.0\%) \\ \hline 
        eXpLogic-L$_{E}$  & \textbf{5.9 $\pm$ 0.3}   &  (91.2\%) & \textbf{3.7 $\pm$ 0.4}   &  (25.4\%)  & \textbf{4.7 $\pm$ 0.5}  &  (80.3\%) \\ 
        eXpLogic-L$_{A}$  & \textbf{5.9 $\pm$ 0.3}   &  (91.6\%) & \textbf{3.7 $\pm$ 0.4}   &  (21.1\%)  & \textbf{4.7 $\pm$ 0.5}  &  (80.4\%) \\ \hline
        eXpLogic-G$_{\mathcal{U}}$  & 7.4 $\pm$ 3.7   &  (92.0\%) & 12.1 $\pm$ 13.2 &  (48.2\%)  & 8.1 $\pm$ 10.2 &  (79.9\%) \\ 
        eXpLogic-G$_{E}$  & 6.6 $\pm$ 7.2   &  (88.7\%) & 14.7 $\pm$ 20.5 &  (7.0\%)   & 5.6 $\pm$ 3.3  &  (86.0\%) \\  
        eXpLogic-G$_{A}$  & 11.1 $\pm$ 6.2  &  (83.3\%) & 11.7 $\pm$ 8.7  &  (57.2\%)  & \textbf{3.7 $\pm$ 2.0}  &  (22.9\%) \\ \hline
        eXpLogic-C$_{E}$  & \textbf{5.6 $\pm$ 1.6}   &  (92.0\%) & 11.2 $\pm$ 12.6 &  (30.2\%)  & 5.1 $\pm$ 2.0  &  (87.0\%) \\ 
        eXpLogic-C$_{A}$  & \textbf{5.5 $\pm$ 1.1}   &  (92.0\%) & 13.2 $\pm$ 14.5 &  (53.9\%)  & 5.3 $\pm$ 5.7  &  (86.2\%) \\ \hline
    \end{tabular}
    \label{tab:switchdist}
\end{table*}

\subsection{Results from MiniNet Model Reduction}
\label{sec:results:inference} 

When reducing the DiffLogic network into MiniNets per predicted class, we generally found the network size (-86\%) and inference time (-10.0\%) were reduced, though the one-vs-all accuracy decreased by 3.8\% on average (see Table \ref{tab:mininet}). The reduced network size indicates that 14\% of a model would need to be loaded into short-term memory (or random access memory) to predict a certain class with reduced inference speeds. These findings may indicate value in leveraging MiniNets in resource-constrained applications, though more work is needed for larger networks and complex datasets.

\begin{table}[!ht]
    \caption{MiniNet comparisons against the parent DiffLogic network, which has 5010 gates, predicts the class score in 337 $\mu$s on average, and predicts classes with a one-versus-all accuracy of 98\% on average. The accuracy (Acc.), the precision (Prec.), recall, and F1 score are reported } % Add a caption if needed
    \scriptsize
    \centering
    \begin{tabular}{c c c c c c c}
    \hline
        Class  & \% Time & \% Size & Acc. & Prec. & Recall & F1 \\ 
        \hline 
        Class 0  & -9.8\% & -85.5\% & 97.2\% & 79.2\% & 11.1\% & 0.87 \\ 
        Class 1  & -10.0\% & -85.6\% & 98.6\% & 89.5\% & 10.9\% & 0.93 \\ 
        Class 2  & -10.1\% & -85.5\% & 94.2\% & 64.7\% & 10.9\% & 0.76 \\ 
        Class 3  & -10.2\% & -85.5\% & 94.0\% & 63.5\% & 11.2\% & 0.76 \\  
        Class 4  & -10.1\% & -85.6\% & 95.0\% & 68.2\% & 11.0\% & 0.79 \\ 
        Class 5  & -10.1\% & -85.5\% & 93.4\% & 61.5\% & 10.8\% & 0.74 \\  
        Class 6  & -10.1\% & -85.5\% & 96.7\% & 76.6\% & 11.1\% & 0.85 \\ 
        Class 7  & -10.1\% & -85.5\% & 96.1\% & 74.0\% & 10.7\% & 0.83 \\ 
        Class 8  & -10.2\% & -85.5\% & 90.4\% & 51.2\% & 10.9\% & 0.65 \\ 
        Class 9  & -10.3\% & -85.5\% & 91.9\% & 55.8\% & 11.1\% & 0.70 \\ 
        \hline 
        Avg. & -10.0\%  & -86.0\%  & 94.8\% & 68.4\% & 11.0\% & 0.789 \\ 
        $\pm$ Std. & $\pm$ 0.1\% & $\pm$ 0.1\% & $\pm$ 2.5\% & $\pm$ 11.5\% & $\pm$ 0.1\% & $\pm$ 0.085 \\         
        \hline 
    \end{tabular}
    \label{tab:mininet}
\end{table}

\section{Conclusions}
\label{sec:conclusion}
Numerous observations emerge from this work. Firstly, eXpLogic provides novel local explanations that help interpret TP, FN, and FP predictions, and reveal whether the absence or presence of certain inputs promoted a model's decision. Secondly, eXpLogic explains what inputs activate certain functions, which can be used to reduce create MiniNets that reduce the resources and inference time needed for class-specific predictions. The eXpLogic saliency maps are generally better than competing methods at identify inputs that change model behaviors, as measured using the SwitchDist metric introduced in this paper. The value to the broader research community is in demonstrating an explanatory approach that explains numerous model behaviors simultaneously, which is valuable in-and-beyond the fields of healthcare, law, and defense. 

\section*{Acknowledgements}
Thanks to the Florida Institute of National Security for sponsoring this work as a part of a Ph.D. dissertation program.

%%\appendix

%%\section{Appendix title 1}
\footnotesize
\bibliography{bibliography}

\begin{thebibliography}{10}

\bibitem{tiwari2023explainable}
Rudra Tiwari.
\newblock Explainable ai (xai) and its applications in building trust and understanding in ai decision making.
\newblock {\em International J. Sci. Res. Eng. Manag}, 7:1--13, 2023.

\bibitem{adadi2018peeking}
Amina Adadi and Mohammed Berrada.
\newblock Peeking inside the black-box: a survey on explainable artificial intelligence (xai).
\newblock {\em IEEE access}, 6:52138--52160, 2018.

\bibitem{petersen2022deep}
Felix Petersen, Christian Borgelt, Hilde Kuehne, and Oliver Deussen.
\newblock Deep differentiable logic gate networks.
\newblock {\em Advances in Neural Information Processing Systems}, 35:2006--2018, 2022.

\bibitem{samuel2021evaluation}
Sam Zabdiel~Sunder Samuel, Vidhya Kamakshi, Namrata Lodhi, and Narayanan~C Krishnan.
\newblock Evaluation of saliency-based explainability method.
\newblock {\em arXiv preprint arXiv:2106.12773}, 2021.

\bibitem{erhan2009visualizing}
Dumitru Erhan, Yoshua Bengio, Aaron Courville, and Pascal Vincent.
\newblock Visualizing higher-layer features of a deep network.
\newblock {\em University of Montreal}, 1341(3):1, 2009.

\bibitem{benamira2022scalable}
Adrien Benamira, Tristan Gu{\'e}rand, Thomas Peyrin, Trevor Yap, and Bryan Hooi.
\newblock A scalable, interpretable, verifiable \& differentiable logic gate convolutional neural network architecture from truth tables.
\newblock {\em arXiv preprint arXiv:2208.08609}, 2022.

\bibitem{vemuri2024enhancing}
Deepika Vemuri, Gautham Bellamkonda, and Vineeth~N Balasubramanian.
\newblock Enhancing concept-based learning with logic.
\newblock In {\em ICML 2024 Next Generation of AI Safety Workshop}.

\bibitem{boccignone2019problems}
Giuseppe Boccignone, Vittorio Cuculo, and Alessandro D’Amelio.
\newblock Problems with saliency maps.
\newblock In {\em Image Analysis and Processing--ICIAP 2019: 20th International Conference, Trento, Italy, September 9--13, 2019, Proceedings, Part II 20}, pages 35--46. Springer, 2019.

\bibitem{adebayo2018sanity}
Julius Adebayo, Justin Gilmer, Michael Muelly, Ian Goodfellow, Moritz Hardt, and Been Kim.
\newblock Sanity checks for saliency maps.
\newblock {\em Advances in neural information processing systems}, 31, 2018.

\bibitem{riche2013saliency}
Nicolas Riche, Matthieu Duvinage, Matei Mancas, Bernard Gosselin, and Thierry Dutoit.
\newblock Saliency and human fixations: State-of-the-art and study of comparison metrics.
\newblock In {\em Proceedings of the IEEE international conference on computer vision}, pages 1153--1160, 2013.

\bibitem{boggust2023saliency}
Angie Boggust, Harini Suresh, Hendrik Strobelt, John Guttag, and Arvind Satyanarayan.
\newblock Saliency cards: A framework to characterize and compare saliency methods.
\newblock In {\em Proceedings of the 2023 ACM Conference on Fairness, Accountability, and Transparency}, pages 285--296, 2023.

\bibitem{tomsett2020sanity}
Richard Tomsett, Dan Harborne, Supriyo Chakraborty, Prudhvi Gurram, and Alun Preece.
\newblock Sanity checks for saliency metrics.
\newblock In {\em Proceedings of the AAAI conference on artificial intelligence}, volume~34, pages 6021--6029, 2020.

\bibitem{jiang2023explaining}
Ziping Jiang.
\newblock On explaining neural network robustness with activation path.
\newblock In {\em The Eleventh International Conference on Learning Representations}, 2023.

\bibitem{yona2021revisiting}
Gal Yona and Daniel Greenfeld.
\newblock Revisiting sanity checks for saliency maps.
\newblock {\em arXiv preprint arXiv:2110.14297}, 2021.

\bibitem{sundararajan2017axiomatic}
Mukund Sundararajan, Ankur Taly, and Qiqi Yan.
\newblock Axiomatic attribution for deep networks.
\newblock In {\em International conference on machine learning}, pages 3319--3328. PMLR, 2017.

\end{thebibliography}



%% For example:

%% \bibitem[Aladro et al.(2015)]{Aladro15} Aladro, R., Martín, S., Riquelme, D., et al. 2015, \aas, 579, A101

%%\end{thebibliography}

\end{document}